\documentclass{article}

\usepackage{graphicx}  

\newcommand{\eat}[1]{}
\usepackage{subcaption} 
\usepackage{booktabs} 
\usepackage{multicol}
\usepackage{multirow}
\usepackage{algorithm,algorithmic}
\usepackage{array}
\usepackage{amsmath,amsfonts}

\newlength{\dhatheight}

\newcommand{\argmax}{\mathop{\rm argmax}}

\newcommand{\ignore}[1]{}
\newcommand{\todo}[1]{}
\newcommand{\oldstuff}[1]{}

\newsavebox{\savepar}

\makeatletter
\newcommand{\vast}{\bBigg@{3}}
\newcommand{\Vast}{\bBigg@{4}}
\makeatother

\usepackage{wrapfig}
\usepackage{enumitem}
\usepackage{nicefrac}       
\usepackage{microtype}      
\usepackage{tabu}
\usepackage{url}
\usepackage{cite}

\allowdisplaybreaks
\newcommand{\bx}{\mathbf{x}}

\newcommand{\cX}{\mathcal{X}}
\newcommand{\Pred}{\textnormal{Pred}}

\newcommand{\bdelta}{\boldsymbol{\delta}}

\makeatletter
\newcommand*{\centerfloat}{%
	\parindent \z@
	\leftskip \z@ \@plus 1fil \@minus \textwidth
	\rightskip\leftskip
	\parfillskip \z@skip}
\makeatother

\title{Leveraging Latent Features for Local Explanations}

\author{Ronny Luss$^1$\thanks{First five authors have equal contribution. 1 and 2 indicate affiliations to IBM Research and University of Michigan respectively.}, Pin-Yu Chen$^{1}$, Amit Dhurandhar$^{1}$, Prasanna Sattigeri$^1$\\ Yunfeng Zhang$^1$, Karthikeyan Shanmugam$^1$, and Chun-Chen Tu$^2$}
\begin{document}
\maketitle




\begin{abstract}
As the application of deep neural networks proliferates in numerous areas such as medical imaging, video surveillance, and self driving cars, the need for explaining the decisions of these models has become a hot research topic, both at the global and local level. Locally, most explanation methods have focused on identifying relevance of features, limiting the types of explanations possible. In this paper, we investigate a new direction by leveraging latent features to generate contrastive explanations; predictions are explained not only by highlighting aspects that are in themselves sufficient to justify the classification, but also by new aspects which if added will change the classification. The key contribution of this paper lies in how we add features to rich data in a formal yet humanly interpretable way that leads to meaningful results. Our new definition of "addition" uses latent features to move beyond the limitations of previous explanations and resolve an open question laid out in Dhurandhar, et. al. (2018), which creates local contrastive explanations but is limited to simple datasets such as grayscale images. The strength of our approach in creating intuitive explanations that are also quantitatively superior to other methods is demonstrated on three diverse image datasets (skin lesions, faces, and fashion apparel). A user study with 200 participants further exemplifies the benefits of contrastive information, which can be viewed as complementary to other state-of-the-art interpretability methods. 
\end{abstract}

\section{Introduction}
With the explosion of deep learning \cite{gan} and its huge impact on domains such as computer vision and speech, amongst others, many of these technologies are being implemented in systems that affect our daily lives. In many cases, a negative side effect of deploying these technologies has been their lack of transparency \cite{lime}, which has raised concerns not just at an individual level \cite{kushtst} but also at an organization or government level\eat{\cite{gdpr}}.

There have been many methods proposed in literature \cite{patternet,Ormas,montavon2017methods,saliency,bach2015pixel} that explain predictions of deep neural networks based on the relevance of different features or pixels/superpixels for an image. Recently, an approach called contrastive explanations method (CEM) \cite{CEM} was proposed which highlights not just correlations or relevances but also features that are minimally sufficient to justify a classification, referred to as pertinent positives (PPs). CEM additionally outputs a minimal set of features, referred to as pertinent negatives (PNs), which when made non-zero or added, alter the classification and thus should remain absent in order for the original classification to prevail. For example, when justifying the classification of a handwritten image of a 3, the method will identify a subset of non-zero or on-pixels within the 3 which by themselves are sufficient for the image to be predicted as a 3 even if all other pixels are turned off (that is, made zero to match background). Moreover, it will identify a minimal set of off-pixels which if turned on (viz. a horizontal line of pixels at the top right making the 3 look like a 5) will alter the classification. Such forms of explanations are not only common in day-to-day social interactions (viz. the twin without the scar) but are also heavily used in fields such as medicine and criminology \cite{pertneg}. \cite{Timcontras} notably surveyed 250 papers in social science and philosophy and found contrastive explanations to be among the most important for human explanations.

To identify PNs, addition is easy to define for grayscale images where a pixel with a value of zero indicates no information and so increasing its value towards 1 indicates addition. However, for color images with rich structure, it is not clear what is a ``no information" value for a pixel and consequently what does one mean by addition. By rich structure, we mean there exists an interpretable latent representation for the data; all faces have a particular shape, hair has a color, and even noses can be described by their shape (e.g., pointy or not).  Defining addition in a naive way such as simply increasing the pixel or red-green-blue (RGB) channel intensities can lead to uninterpretable images as the relative structure may not be maintained with the added portion not necessarily being interpretable. Moreover, even for grayscaled images just increasing values of pixels may not lead to humanly interpretable images nor is there a guarantee that the added portion can be interpreted even if the overall image is realistic and lies on the data manifold.

In this paper, we overcome these limitations by defining ``addition" in a novel way which leads to realistic images with the additions also being interpretable. This work is an important contribution toward explaining black-box model predictions because it is applicable to a large class of data sets (whereas \cite{CEM} was very limited in scope and cannot explain data using the rich structure described above). To showcase the general applicability of our method to various settings, we first experiment with ISIC Lesions \cite{isic1,isic2} where the data manifold is learned using a variational autoencoder (VAE) \cite{dipvae} and certain (interpretable) latent factors (as no attributes are available) are used to create realistic images with interpretable additions. Our second experiment is on CelebA \cite{celebA} where we apply our method to a data manifold learned using a generative adversarial network (GAN) \cite{gan} trained over the data and by building attribute classifiers for certain high-level concepts (viz. lipstick, hair color) in the dataset. We create realistic images with interpretable additions. 

While our contributions are motivated by and related to CEM of \cite{CEM}, the ISIC Lesions and CelebA datasets are color images and hence models classifying them cannot be explained by CEM. We thus compare against other state-of-the-art methods described in the Experiments section. A third experiment with the grayscale image dataset Fashion-MNIST \cite{fashionmnist} that compares with CEM and maintains the benefits of our work is included for completeness.

Please note our assumption that high-level latent interpretable features can be learned if not given is not the subject of this paper and has been addressed in multiple existing works, e.g., \cite{infogan, tcav}. A key benefit to disentangled representations is interpretability as discussed in \cite{dipvae}. The practicality and interpretability of disentangled features is validated here by its use in the ISIC Lesions and Fashion-MNIST datasets.

The three usecases show that our method can be applied to color as well as grayscale images and to datasets that may or may not have high level attributes. It is important to note that our goal is not to generate interpretable models that are understood by humans, but rather to explain why a fixed model outputs a particular prediction. We conducted a human study which concludes that our method offers superior explanations compared to other methods, in particular because our contrastive explanations go beyond visual explanations, which is key to human comprehension. \eat{Moreover, for the PPs, we are able to select subsets of the image, i.e. superpixels that are sufficient to maintain the original classification, with the PPs being sparse both in the number of superpixels as well as the number of high-level semantic concepts they represent.}

\vspace{-0.1cm}
\section{Related Work}

There have been many methods proposed in literature that aim to explain reasons for their decisions. These methods may be globally interpretable -- rule/decision lists \cite{decl,twl}, or exemplar based -- prototype exploration \cite{l2c,proto}, or inspired by psychometrics \cite{irt} or interpretable generalizations of  linear models \cite{Caruana:2015}. There are also works that try to formalize interpretability \cite{tip}. 

A survey by \cite{montavon2017methods} mainly explores two methods for explaining decisions of neural networks: i) Prototype selection methods \cite{nguyen2016synthesizing,nguyen2016multifaceted} that produce a prototype for a given class
, and ii) Local explanation methods that highlight relevant input features/pixels \cite{bach2015pixel,patternet,lime,unifiedPI}. 
Belonging to this second type are multiple local explanation methods that generate explanations for images \cite{gradcam,saliency,Ormas, fong17, fong19} and some others for NLP applications \cite{lei2016rationalizing}. There are also works \cite{tcav} that highlight higher level concepts present in  images based on examples of concepts provided by the user. These methods mostly focus on features that are present, although they may highlight negatively contributing features to the final classification. In particular, they do not identify concepts or features that are minimally sufficient to justify the classification or those that should be necessarily absent to maintain the original classification. There are also methods that perturb the input and remove features \cite{QEval} to verify their importance, but these methods can only evaluate an explanation and do not find one.

Recently, there have been works that look beyond relevance. In \cite{anchors}, the authors try to find features that, with almost certainty, indicate a particular class. These can be seen as 
global indicators for a particular class. Of course, these may not always exist for a dataset. There are also works \cite{symbolic} that try to find stable insight that can be conveyed to the user in a (asymmetric) binary setting for medium-sized neural networks. A recent work for generating contrastive explanations \cite{grace} was designed specifically for tabular data and is not applicable here. The most relevant work to our method is \cite{CEM} which cannot be applied when it comes to explaining color images or images with rich structure.
\vspace{-0.05in}
\section{Methodology}

We now describe our Contrastive Explanations Method using Monotonic Attribute Functions (CEM-MAF) for generating contrastive explanations for images and provide algorithmic details for solving the optimization problems in Algorithm \ref{cemmaf}. An implementation is available as part of \url{https://github.com/Trusted-AI/AIX360}. \eat{ We now describe how to identify PNs, which involves the key contribution of defining ``addition" for color images in a meaningful way. We then describe how to identify PPs, which also utilizes this notion of adding attributes. Finally, we provide algorithmic details for solving the optimization problems in Algorithm \ref{cemmaf}.}

Given $k$ (available or learned) interpretable features (latent or otherwise) which represent meaningful concepts (viz. moustache, glasses, smile), let $g_i(.), \forall i\in \{1, ..., k\},$ be corresponding functions acting on these features with higher values indicating presence of a certain visual concept while lower values indicating its absence. For example, CelebA has different high-level (interpretable) features for each image such as whether the person has black hair or high cheekbones. In this case, we build binary classifiers for each of the features where a 1 indicates presence of black hair or high cheekbones, while a 0 indicates its absence. These classifiers would be the $g_i(.)$ functions. On the other hand, for datasets with no high-level features, we find latents by learning disentangled representations and choose those latents that are interpretable. Here the $g_i$ functions would be an identity (or negative identity) map depending on which direction adds a certain concept (viz. light to dark colored lesion). We note that these attribute functions could be used as latent features for the generator in a causal graph (e.g., \cite{causalgan}), or given a causal graph for the desired attributes, we could learn these functions from the architecture in \cite{causalgan}. 
 
Our procedure for finding PNs and PPs involves solving optimization problems over the variable $\bdelta$ which is the output image. As discussed below, in the case of PNs, $\bdelta$ is the output of a GAN or VAE that generates a new image, while in the case of PPs, $\bdelta$ is the result of applying a mask to the original input. While $\bdelta$ for PNs and PPs lie in different spaces, we maintain that both PNs and PPs should be discussed together because they offer complementary information towards explaining a classification model.   

In what follows, $\cX$ denotes the feasible input space with $(\bx_0,t_0)$ an example such that $\bx_0 \in \cX$ and $t_0$ is the predicted label obtained from a classifier $f(\cdot)$, where $f(\cdot)$ is any function that outputs a vector of prediction scores for all classes.\eat{, such as prediction probabilities or logits (unnormalized probabilities) that are widely used in neural networks.} 

\begin{algorithm}[tb]
    \caption{Contrastive Explanations Method using Monotonic Attribute Functions (CEM-MAF)}
    \label{cemmaf}
\begin{algorithmic}
\STATE \textbf{Input:} Example $(\bx_0,t_0)$, latent representation of $\bx_0$ denoted $z_{\bx_0}$, neural network classification model $f(.)$, set of (binary) masks $\mathcal{M}$, and $k$ monotonic feature functions $g=\{g_1(.), ..., g_k(.)\}$.\\
\STATE 1) Solve (\ref{eqn_per_neg}) on example $(\bx_0,t_0)$ and obtain $\bdelta^{\textnormal{PN}}$ as the minimizing Pertinent Negative.
\STATE 2) Solve (\ref{eqn_per_pos}) on example $(\bx_0,t_0)$ and obtain $\bdelta^{\textnormal{PP}}$ as the minimizing Pertinent Positive. 
\STATE return $\bdelta^{\textnormal{PN}}$ and $\bdelta^{\textnormal{PP}}$.
\end{algorithmic}
\end{algorithm}

\subsection{Pertinent Negatives (PNs)}

To find PNs, we want to create a (realistic) image that lies in a different class than the original image but where we can claim that we have minimally ``added" things to the original image without deleting anything to obtain this new image. If we are able to do this, we can say that the additions, which we call PNs, should be necessarily absent from the original image in order for its classification to remain unchanged.

The question is how to define ``addition" for color images with rich structure. In \cite{CEM}, the authors tested on grayscale images where intuitively it is easy to define addition as increasing the pixel values towards 1. This, however, does not generalize to images where there are multiple channels (viz. RGB) and inherent structure that leads to realistic images, and where simply moving away from a certain pixel value may lead to unrealistic images. Moreover, addition defined in this manner may be completely uninterpretable. This is true even for grayscale images where, while the final image may be realistic, the addition may not be. The other big issue is that for color images, the background may be any color that indicates no signal; object pixels with lower value in the original image, if increased towards background, will make the object imperceptible in the new image, while the claim would be that we have ``added" something. Such counterintuitive cases arise for complex images if we maintain the definition of addition in \cite{CEM}.

Given these issues, we define addition in a novel manner. To define addition we assume that we have high-level interpretable features available for the dataset. Multiple public datasets \cite{celebA,deepfashion} have high-level interpretable features, while for others such features can be learned using unsupervised methods such as disentangled representations \cite{dipvae} or supervised methods where one learns concepts through labeled examples \cite{tcav}. Given $k$ such features, we define functions $g_i(.), \forall i\in\{1, ..., k\},$ as above, where in each of these functions, increasing value indicates addition of a concept. Using these functions, we define addition as introducing more concepts into an image without deleting existing concepts. Formally, this corresponds to never decreasing the $g_i(.)$ from their original values based on the input image, but rather increasing them. However, we also want a minimal number of additions for our explanation to be interpretable and so we encourage as few $g_i()$ as possible to increase in value (within their allowed ranges) that will result in the final image being in a different class. 

We also want the final image to be realistic and thus we first learn a data manifold, denoted $\mathcal{D}(\cdot)$, using a GAN or VAE on which we can perturb the image, as we want our final image to also lie on it after the necessary additions. Let $z$ denote the latent representation with $z_{x}$ denoting the latent representation corresponding to input $\bx$ such that $\bx=\mathcal{D}(z_{x})$. This gives rise to the following optimization problem:
\begin{align}
\label{eqn_per_neg}
\min_{\bdelta \in \cX}~&\gamma \sum_i\max\{ g_i(\bx_0)-g_i(\mathcal{D}(z_{\bdelta})),0\} + \beta \|g\left(\mathcal{D}(z_{\bdelta})\right)\|_1\nonumber\\& 
- c \cdot \min \{ \max_{i \neq t_0} [f(\bdelta)]_i - [f(\bdelta)]_{t_0}, \kappa   \}\nonumber\\& +\eta \|\bx_0-\mathcal{D}(z_{\bdelta})\|_2^2+\nu \|z_{\bx_0}-z_{\bdelta}\|_2^2.
\end{align}
The first two terms in the objective function here are the novelty for PNs. The first term encourages the addition of attributes where we want the $g_i(.)$s for the final image to be no less than their original values. The second term encourages minimal addition of interpretable attributes. The third term is the PN loss from \cite{CEM} and encourages the modified example $\bdelta$ to be predicted as a different class than $t_0=\arg \max_i [f(\bx_0)]_i$, where $[f(\bdelta)]_i$ is the $i$-th class prediction score of $\bdelta$. The hinge-like loss function pushes the modified example $\bdelta$ to lie in a different class than $\bx_0$. The parameter $\kappa \geq 0$ is a confidence parameter that controls the separation between $[f(\bdelta)]_{t_0}$ and  $\max_{i \neq t_0} [f(\bdelta)]_i$. The fourth ($\eta>0$) and fifth ($\nu>0$) terms encourage the final image to be close to the original image in the input and latent spaces, respectively. In practice, one could have a threshold for each of the $g_i(.)$, where only an increase in values beyond that threshold would imply a meaningful addition. The advantage of defining addition in this manner is that not only are the final images interpretable, but so are the additions, and we can clearly elucidate which (concepts) should be necessarily absent to maintain the original classification. Finally, we note that formulation (\ref{eqn_per_neg}) could be equivalently written as an optimization problem over the latent space since $\delta=\mathcal{D}(z_{\delta})$, and in fact, our implementation does optimize over the latent variable $z_{\delta}$; however, optimizing over an image, viz. (\ref{eqn_per_neg}), is easier to understand.

\subsection{Pertinent Positives (PPs)}

To find PPs, we want to highlight a minimal set of important pixels or superpixels (from a segmentation of input $\bx_0$) which by themselves are sufficient for the classifier to output the same class as the original example. Let $\mathcal{M}$ denote a set of binary masks which when applied to $\bx_0$ produce images $\mathcal{M}(x_0)$ by selecting the corresponding superpixels from the segmentation of $\bx_0$. Let $M_x$ denote the mask corresponding to image $\bx=M_x(\bx_0)$ when applied on input $\bx_0$.

Our goal, for example image $\bx_0$, is to find an image $\bdelta\in \mathcal{M}(x_0)$ such that $\argmax_i [\Pred(\bx_0)]_i = \argmax_i [\Pred(\bdelta)]_i$ (i.e. same prediction), with $\bdelta$ containing as few superpixels and interpretable concepts from the original image as possible. This leads to the following optimization problem:
\begin{align}
\label{eqn_per_pos}
\min_{\bdelta \in \mathcal{M}(\bx_0)}~&\gamma \sum_i \max\{g_i\left(\bdelta\right)-g_i(\bx_0),0\} +  \beta \|M_{\bdelta}\|_1\nonumber\\& - c \cdot \min \{[f(\bdelta)]_{t_0} - \max_{i \neq t_0} [f(\bdelta)]_i, \kappa   \}.
\end{align}
The first term in the objective function here is the novelty for PPs and penalizes the addition of attributes since we seek a sparse explanation. The last term is the PP loss from \cite{CEM} and is minimized when  $[f(\bdelta)]_{t_0}$ is greater than $\max_{i \neq t_0} [f(\bdelta)]_i$ by at least $\kappa\geq 0$, which is a margin/confidence parameter. Parameters $\gamma, c, \beta \geq 0$ are the associated regularization coefficients.

In the above formulation, we optimize over superpixels which of course subsumes the case of just using pixels. Superpixels have been used in prior works \cite{lime} to provide more interpretable results on image datasets and we allow for this more general option.

\subsection{Optimization Details}
\label{ad}
To solve for PNs as formulated in \eqref{eqn_per_neg}, we note that the $L_1$ regularization term is penalizing a non-identity and complicated function $\|g\left(\mathcal{D}(z_{\bdelta})\right)\|_1$ of the optimization variable $\bdelta$ involving the data manifold $\mathcal{D}$, so proximal methods are not applicable. Instead, we use 1000 iterations of standard subgradient descent to solve \eqref{eqn_per_neg}. We find a PN by setting it to be the iterate having the smallest $L_2$ distance $\|z_{\bx_0}-z_{\bdelta}\|_2$ to the latent code $z_{\bx_0}$ of $\bx_0$, among all iterates where the prediction score of solution $\bdelta^*$ is at least $[f(\bx_o)]_{t_0}$.

To solve for PPs as formulated in \eqref{eqn_per_pos}, we first relax the binary mask  $M_{\bdelta}$ on superpixels to be real-valued (each entry is between $[0,1]$) and then apply the standard iterative soft thresholding algorithm (ISTA) (see \cite{beck2009fast} for various references) that efficiently solves optimization problems with $L_1$ regularization. We run 1000 iterations of ISTA in our experiments and obtain a solution $M_{\bdelta^*}$ that has the smallest $L_1$ norm and satisfies the prediction of $\bdelta^*$ being within margin $\kappa$ of $[f(\bx_o)]_{t_0}$. We then rank the entries in $M_{\bdelta^*}$ according to their values in descending order and subsequently add ranked superpixels until the masked image predicts $t_0 = \argmax_i[f(\bx_o)]_{i}$.

While several parameters, $\beta, c,\gamma, \nu$ and $\eta$, require tuning, it is important to consider that for explainability, in practice, one would only tune these models once over a validation set, after which CEM-MAF can be used to explain an arbitrary number of future predictions by the model. Further discussion of hyperparameter selection for replication purposes is held to the Supplement.
\vspace{-.05in}

\section{Experiments}
We next illustrate the usefulness of CEM-MAF on three image datasets - ISIC Lesions \cite{isic1,isic2}, CelebA \cite{celebA}, and Fashion-MNIST \cite{fashionmnist}. These datasets cover the gamut of images having known versus derived features and color versus grayscale. CEM-MAF handles each scenario and offers explanations that are understandable by humans. We include Fashion-MNIST in our study solely in order to demonstrate the advantages of CEM-MAF over CEM \cite{CEM}; as previously noted, CEM is only amenable to grayscale images and it was particularly this limitation of CEM that motivated the use of latent features to derive a new framework for contrastive explanations that is amenable to color images such as ISIC Lesions and CelebA.

We make the following observations:
\setlist[itemize]{leftmargin=*}
\begin{itemize}
\item PNs offer intuitive explanations given a set of interpretable monotonic attribute functions. In fact, our user study finds them to be the preferred form of explanation; describing a decision in isolation (viz. why is a smile a smile) using PPs, relevance, or heatmaps is not always informative.
\item PPs offer better direction as to what is important for the classification versus too much direction by LIME (shows too many features) or too little direction by Grad-CAM (only focuses on smiles). 
\item PPs and PNs, by definition, offer the guarantee of being 100\% accurate in maintaining or changing the class respectively as seen in Table 1 versus LIME or Grad-CAM.
\eat{\item Both proximity in the input and latent space along with sparsity in the additions play an important role in generating good quality human-understandable contrastive explanations.}
\end{itemize}

\noindent\textbf{PN vs PP: } PNs, as shown below, offer directly actionable explanations for the classifier - if the stated changes were made to the data point, the classifier would change its prediction. PPs are actionable in the sense that if the user deems the selected features to be irrelevant, then the classifier should be further investigated. In this regard, they can be used in similar fashion to LIME and Grad-CAM, but we also demonstrate below the benefits of PPs quantitatively in terms of feature selection. Note that PPs offer information according to a specific definition - it is out of the scope of this work to determine how to act on this information.

\noindent\textbf{Comparative Explanation Methods: } PNs and PPs are compared below with two methods: one derived from locally interpretable model-agnostic explanations (LIME) \cite{lime} and a gradient-based localized explanation designed for CNN models (Grad-CAM) \cite{gradcam}. As noted in the Related Work section, there are many possible explanation tools to compare with, but LIME and Grad-CAM (or one of various related gradient-based methods that followed) are the two most commonly compared with in most related literature. Regarding our close connection to CEM \cite{CEM}, recall that our goal is to explain models for which CEM is not applicable. CEM is not applicable to models that classify ISIC Lesions and CelebA because it cannot handle color images. However, to complete the story, a third experiment with Fashion-MNIST \cite{fashionmnist} compares explanations from CEM with CEM-MAF and demonstrates that, even on grayscale images, there is a benefit to using high-level features for explanations. 
 
 \begin{figure}[htbp]
 	\center \small
 	\begin{tabu}{|c|c|c|c|c|c|c|}
 		\hline
 		\raisebox{0.5ex}{\textbf{\shortstack{Orig. \\ Class}}}
 		& \raisebox{1.0ex}{\shortstack{Mela \\ -noma}}
 		& \raisebox{1.0ex}{\shortstack{B. Kera \\ -tosis}}
 		& \raisebox{1.0ex}{\shortstack{B. Cell \\ Carc.}}
 		& \raisebox{1.0ex}{\shortstack{Vasc. \\ Lesion}}
 		& \raisebox{1.0ex}{\shortstack{Vasc. \\ Lesion}}
 		& \raisebox{1.5ex}{Nevus}
 		\\ \hline
 		
 		\raisebox{3.0ex}{\textbf{Orig.}}
 		&\includegraphics[width=.05\textwidth]{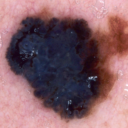}
 		& \includegraphics[width=.05\textwidth]{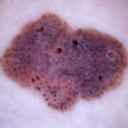}
 		& \includegraphics[width=.05\textwidth]{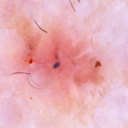}
 		& \includegraphics[width=.05\textwidth]{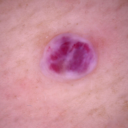}
 		& \includegraphics[width=.05\textwidth]{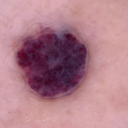}
 		& \includegraphics[width=.05\textwidth]{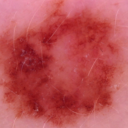}
 		\\ \hline
 		
 		\raisebox{0.5ex}{\textbf{\shortstack{PN \\ Class}}}
 		& \raisebox{1.0ex}{Nevus}
 		& \raisebox{0.5ex}{\shortstack{Mela \\ -noma}}
 		& \raisebox{1.5ex}{Nevus}
 		& \raisebox{0.5ex}{\shortstack{Bas. Cell \\ Carc.}}
 		& \raisebox{1.5ex}{Nevus}
 		& \raisebox{1.0ex}{\shortstack{Vasc. \\ Lesion}}
 		\\ \hline
 		
 		\raisebox{3.0ex}{\textbf{PN}}
 		& \includegraphics[width=.05\textwidth]{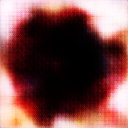}
 		&\includegraphics[width=.05\textwidth]{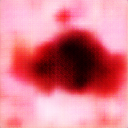}
 		&\includegraphics[width=.05\textwidth]{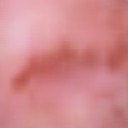}
 		&\includegraphics[width=.05\textwidth]{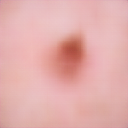}
 		&\includegraphics[width=.05\textwidth]{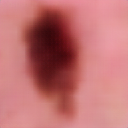}
 		&\includegraphics[width=.05\textwidth]{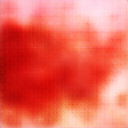}
 		\\ \hline
 		
 		\raisebox{0.5ex}{\textbf{\shortstack{PN \\ Expl.}}}
 		&\raisebox{2.0ex}{rounder}
 		&\raisebox{2.0ex}{darker}
 		&\raisebox{1.0ex}{\shortstack{rounder, \\lighter}}
 		&\raisebox{1.0ex}{\shortstack{more \\oval, \\ smaller}}
 		&\raisebox{1.0ex}{\shortstack{more \\defined \\ border}}
 		&\raisebox{1.0ex}{\shortstack{rounder, \\smaller}}
 		\\ \hline
 		
 		\raisebox{1.5ex}{\textbf{PP}}
 		& \includegraphics[width=.05\textwidth]{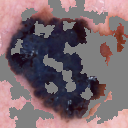}
 		&\includegraphics[width=.05\textwidth]{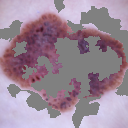}
 		&\includegraphics[width=.05\textwidth]{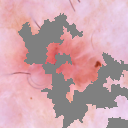}
 		&\includegraphics[width=.05\textwidth]{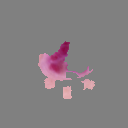}
 		&\includegraphics[width=.05\textwidth]{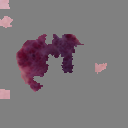}
 		&\includegraphics[width=.05\textwidth]{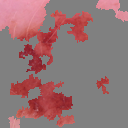}
 		\\ \hline
 		
 		\raisebox{3.0ex}{\textbf{LIME}}
 		& \includegraphics[width=.05\textwidth]{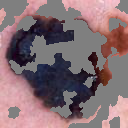}
 		&\includegraphics[width=.05\textwidth]{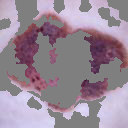}
 		&\includegraphics[width=.05\textwidth]{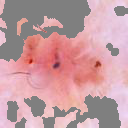}
 		&\includegraphics[width=.05\textwidth]{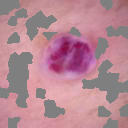}
 		&\includegraphics[width=.05\textwidth]{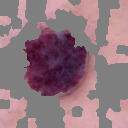}
 		&\includegraphics[width=.05\textwidth]{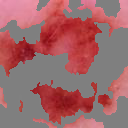}
 		\\ \hline
 		
 		\raisebox{1.5ex}{\textbf{\shortstack{Grad \\-CAM}}}
 		&\includegraphics[width=.05\textwidth]{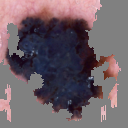}
 		&\includegraphics[width=.05\textwidth]{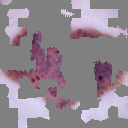}
 		&\includegraphics[width=.05\textwidth]{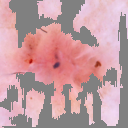}
 		&\includegraphics[width=.05\textwidth]{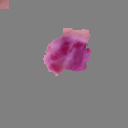}
 		&\includegraphics[width=.05\textwidth]{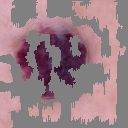}
 		&\includegraphics[width=.05\textwidth]{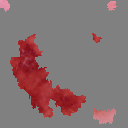}
 		\\ \hline
 		\multicolumn{7}{c}{} \\
 		\multicolumn{7}{c}{\footnotesize{(a) ISIC Lesion dataset}} \\
 		\multicolumn{7}{c}{} \\
 		\multicolumn{7}{c}{} \\
 		\hline
 		\raisebox{0.5ex}{\textbf{\shortstack{Orig. \\ Class}}}
 		& \raisebox{1.0ex}{\shortstack{yng, ml, \\ not smlg}}
 		& \raisebox{1.0ex}{\shortstack{yng, fml, \\ smlg}}  
 		& \raisebox{1.0ex}{\shortstack{yng, fml, \\ not smlg}}
 		& \raisebox{1.0ex}{\shortstack{old, ml, \\ not smlg}}
 		& \raisebox{1.0ex}{\shortstack{yng, ml, \\ smlg}}  
 		& \raisebox{1.0ex}{\shortstack{yng, ml, \\ smlg}}  
 		\\ \hline
 		
 		\raisebox{3.0ex}{\textbf{Orig.}}
 		& \includegraphics[width=.05\textwidth]{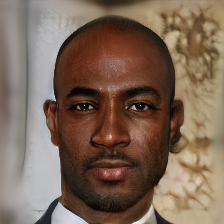}
 		& \includegraphics[width=.05\textwidth]{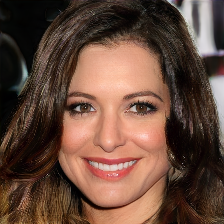}
 		& \includegraphics[width=.05\textwidth]{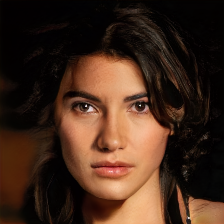}
 		& \includegraphics[width=.05\textwidth]{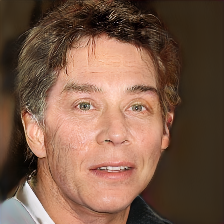}
 		& \includegraphics[width=.05\textwidth]{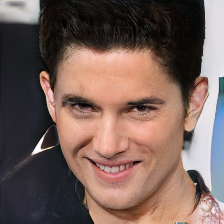}
 		& \includegraphics[width=.05\textwidth]{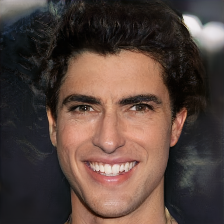}
 		\\ \hline
 		
 		\raisebox{0.5ex}{\textbf{\shortstack{PN \\ Class}}}
 		& \raisebox{0.5ex}{\shortstack{yng, ml, \\ \textbf{smlg}}}
 		& \raisebox{0.5ex}{\shortstack{\textbf{old}, fml, \\ smlg}}  
 		& \raisebox{0.5ex}{\shortstack{yng, \textbf{ml}, \\ not smlg}}
 		& \raisebox{0.5ex}{\shortstack{old, ml, \\ \textbf{smlg}}}
 		& \raisebox{1.0ex}{\shortstack{yng, \textbf{fml}, \\ smlg}}  
 		& \raisebox{1.0ex}{\shortstack{yng, \textbf{fml}, \\ smlg}}  
 		\\ \hline
 		
 		\raisebox{3.0ex}{\textbf{PN}}
 		& \includegraphics[width=.05\textwidth]{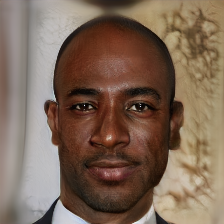}
 		& \includegraphics[width=.05\textwidth]{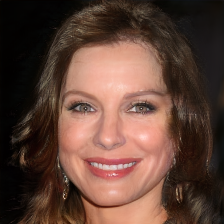}
 		& \includegraphics[width=.05\textwidth]{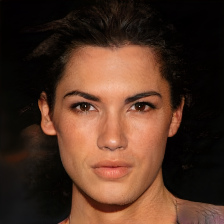}
 		& \includegraphics[width=.05\textwidth]{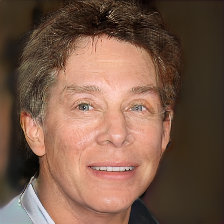}
 		& \includegraphics[width=.05\textwidth]{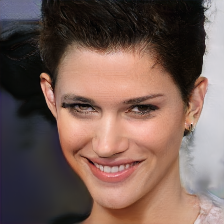}
 		& \includegraphics[width=.05\textwidth]{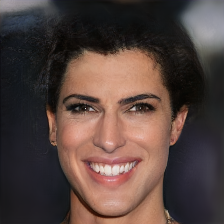}
 		\\ \hline
 		
 		\raisebox{0.5ex}{\textbf{\shortstack{PN \\ Expl.}}}
 		& \raisebox{1.5ex}{\shortstack{+cheek \\ -bones}}
 		& \raisebox{1.5ex}{\shortstack{+oval \\ face}}  
 		& \raisebox{0.5ex}{\shortstack{+single \\ hair clr, \\ -bangs}}
 		& \raisebox{1.5ex}{\shortstack{+cheek \\ -bones}}
 		& \raisebox{1.0ex}{\shortstack{+makeup \\ +black \\ hair}}
 		& \raisebox{2.5ex}{+makeup}
 		\\ \hline
 		
 		\raisebox{1.5ex}{\textbf{PP}}
 		& \includegraphics[width=.05\textwidth]{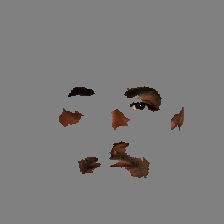}
 		& \includegraphics[width=.05\textwidth]{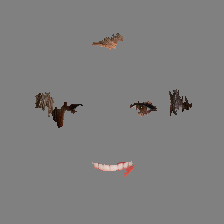}
 		& \includegraphics[width=.05\textwidth]{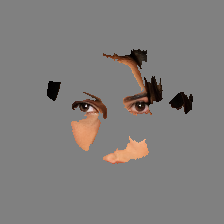}
 		& \includegraphics[width=.05\textwidth]{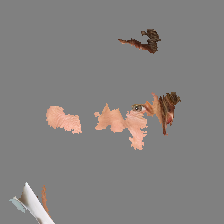}
 		& \includegraphics[width=.05\textwidth]{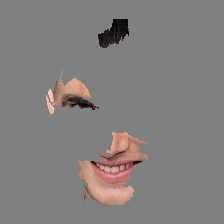}
 		& \includegraphics[width=.05\textwidth]{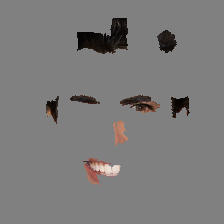}
 		\\ \hline
 		
 		\raisebox{3.0ex}{\textbf{LIME}}
 		& \includegraphics[width=.05\textwidth]{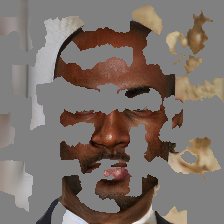}
 		& \includegraphics[width=.05\textwidth]{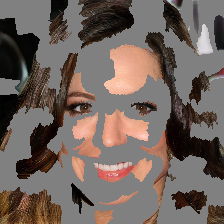}
 		& \includegraphics[width=.05\textwidth]{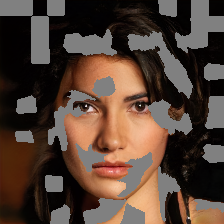}
 		& \includegraphics[width=.05\textwidth]{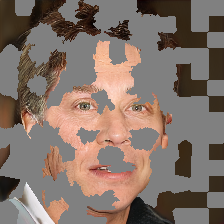}
 		& \includegraphics[width=.05\textwidth]{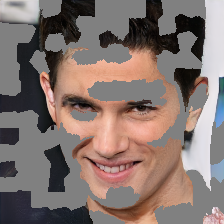}
 		& \includegraphics[width=.05\textwidth]{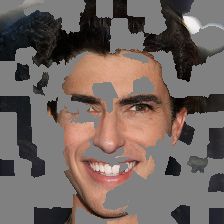}
 		\\ \hline
 		
 		\raisebox{1.5ex}{\textbf{\shortstack{Grad \\-CAM}}}
 		& \includegraphics[width=.05\textwidth]{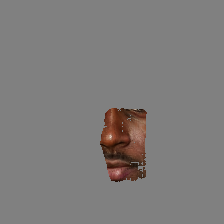}
 		& \includegraphics[width=.05\textwidth]{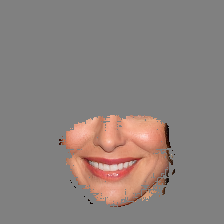}
 		& \includegraphics[width=.05\textwidth]{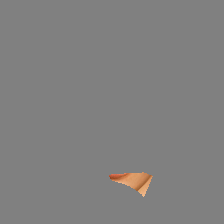}
 		& \includegraphics[width=.05\textwidth]{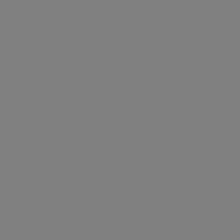} 
 		& \includegraphics[width=.05\textwidth]{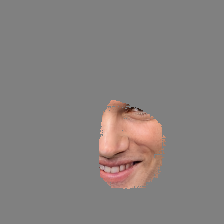} 
 		& \includegraphics[width=.05\textwidth]{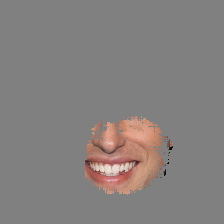} 
 		\\ \hline
 		\multicolumn{7}{c}{} \\
 		\multicolumn{7}{c}{\footnotesize{(b) CelebA dataset}}
 	\end{tabu}
 	
 	\caption{CEM-MAF examples on (a) ISIC Lesions and (b) CelebA, both using segmentation with 200 superpixels. Abbreviations are: Orig. for Original, PN for Pertinent Negative, PP for Pertinent Positive, and Expl. for Explanation. In (a), change from original prediction is based on change in disentangled features. In (b), change from original class prediction is in bold in PN class prediction. Abbreviations in class predictions are as follows: yng for young, ml for male, fml for female, smlg for smiling.}
 	\label{fig:celeba}
 \end{figure}
 
\subsection{Skin Lesions with Disentangled Features} 
\begin{figure}[t]
\begin{center}
  \begin{tabular}{cc}
 Increasing Size & Circular to Oval Shape\\
   \includegraphics[width=.45\textwidth]{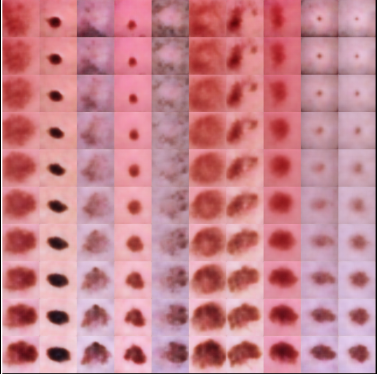} &
    \includegraphics[width=.45\textwidth]{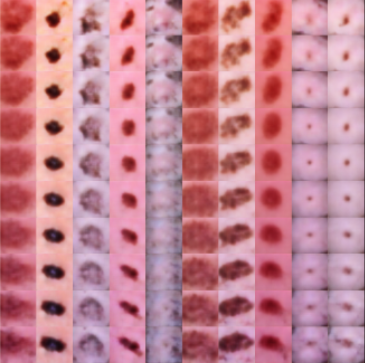}
        \end{tabular}
  \caption{Feature disentanglement of ISIC Lesions images. Both images represent one-dimensional features that vary in [-3, 3]. As the feature shown in the left figure increases from -3 to 3, a corresponding lesion would increase in size. As the feature shown in the right figure increases from -3 to 3, a lesion would change from circular to more oval.}
    \label{fig:disentangled_features}
\end{center}
\vskip  -.2in
\end{figure}

The International Skin Imaging Collaboration (ISIC) Archive (\url{https://challenge2018.isic-archive.com/}) consists of dermoscopic (photographic technique for removing surface reflections) images of skin lesions \cite{isic1,isic2} that fall into one of seven categories.\eat{ Descriptions of seven different types of lesions can be found in the Supplement.} For ISIC Lesions, we explain a 7-class classifier trained on a Resnet9 architecture (classes such as melanoma, nevus, etc.). We train a VAE to generate realistic images for PNs. This data does not have annotated features as in the following example with CelebA, so we learn features using disentanglement following a recent variant of VAE called DIP-VAE \cite{dipvae}\eat{(see Supplement for more details)}. We can then use these disentangled latent features in lieu of the ground truth attributes. Based on what might be relevant to the categories of skin lesions, we use ten dimensions from the latent code as attributes, corresponding to size, roundness, darkness, and border definition. Figure \ref{fig:disentangled_features} visualizes two features corresponding to increasing size and shape changing from circular to oval as the features increase.  \eat{A visualization of all disentangled features, including for Fashion-MNIST, can be found in the Supplement.}

\subsubsection{ISIC Lesions Observations}
Results on six images are exhibited in Figure \ref{fig:celeba}(a) using a segmentation into 200 superpixels\eat{ (more examples in  Supplement)}. The first two rows show the original class prediction followed by the original image. The next two rows show the pertinent negative's class prediction and the pertinent negative image. The fifth row lists the attributes that were modified in the original, i.e., the reasons why the original image is not classified as being in the class of the pertinent negative. The next row shows the pertinent positive image, which combined with the PN, gives the complete explanation. The final two rows show the LIME and Grad-CAM explanations. 

Explanations for the PPs can be determined using descriptions of the lesions. The first PP for an image of Melanoma leaves parts of the lesion displaying two different colors, black and brown (these lesions usually do not have a single solid color). The third PP of Basal Cell Carcinoma highlights the pink skin color associated with the most common type of such lesions (and removes the border meaning the prediction does not focus on the border). LIME highlights similar areas, while using more superpixels. Grad-CAM mostly does not highlight specific areas of the lesions, but rather the entire lesion. Grad-CAM removes the brown part of the first Melanoma and the resulting prediction is incorrect (Melanoma usually does not have a single solid color). \eat{Note that Nevus lesion PPs contain a single superpixel due to heavy dataset bias for this class.}

\begin{table}[t]
\begin{center}
\caption{Quantitative comparison of CEM-MAF, LIME, and Grad-CAM. \# PP Features and PP Accuracy shows that CEM-MAF strictly maintains the original class while using less features than LIME and Grad-CAM. PP Correlation shows that CEM-MAF is best at selecting features that increase confidence in the classifier's prediction on color images and on par with LIME on grayscale images that are easier to classify. At least 100 samples is used per dataset to compute statistics. 
}
\begin{tabular}{|c|c|c|c|c|}
\hline
& & \textbf{\# PP} & \textbf{PP} & \textbf{PP Corr}\\
\textbf{Dataset}  & \textbf{Method} &\textbf{Features} & \textbf{Accuracy} & \textbf{-elation} \\ \hline
\multirow{3}{*}{\shortstack{ISIC \\ Lesions}} & CEM-MAF &  25.8 $(\pm 1.9)$& 100  $(\pm 0.0)$& -.73 $(\pm .04)$\\\cline{2-5} 
& LIME & 79.3 $(\pm 0.9)$  & 98.9 $(\pm 0.5)$ & -.61 $(\pm .04)$  \\\cline{2-5} 
& Grad-CAM & 174 $(\pm 1.5)$  & 97.4 $(\pm 0.7)$  & -.42 $(\pm .06)$\\\hline
\multirow{3}{*}{CelebA} & CEM-MAF & 11.1 $(\pm 0.6)$ & 100  $(\pm 0.0)$& -.57 $(\pm .04)$ \\\cline{2-5} 
& LIME & 66.0 $(\pm 4.9)$ & 56.8 $(\pm 5.1)$& -.41 $(\pm .06)$  \\\cline{2-5} 
& Grad-CAM & 32.0 $(\pm 4.0)$ & 32.6 $(\pm 4.8)$ & .11 $(\pm .01)$ \\\hline
Fashion & CEM-MAF & 52.1 $(\pm 2.0)$ &  100 $(\pm 0.0)$ & -.50 $(\pm .04)$ \\\cline{2-5} 
MNIST& LIME & 236 $(\pm 6.4)$ & 95.1 $(\pm 2.2)$ & -.47 $(\pm .05)$ \\\cline{2-5} 
Clothes& Grad-CAM & 330 $(\pm 13)$ & 58.0 $(\pm 5.1)$ & .19 $(\pm .06)$ \\\hline
Fashion & CEM-MAF &38.9 $(\pm 2.5)$ & 100 $(\pm 0.0)$ & -.41 $(\pm .06)$ \\\cline{2-5} 
MNIST & LIME &240 $(\pm 6.8)$ & 100 $(\pm 6.8)$ & -.58 $ (\pm .04)$ \\\cline{2-5}
Shoes& Grad-CAM & 210 $(\pm 9.5)$ & 78.8 $(\pm 4.2)$ & -.27 $(\pm .06)$ \\\hline

\end{tabular}
\label{tab:performance}
\end{center}
\vspace{-.1in}
\end{table}

\eat{
\begin{wraptable}{r}{0.35\textwidth} 
\centering
\caption{Quantitative comparison of CEM-MAF, LIME, and Grad-CAM on  CelebA and Lesions. }
\begin{tabular}{|c|c|c|}
\hline
\textbf{PP}&\textbf{PP} & \textbf{PP}\\
\textbf{Dataset}  & \textbf{Method} & \textbf{Corr} \\ \hline
\multirow{3}{*}{CelebA} & CEM-MAF & -.742 \\\cline{2-3} 
& LIME & .035  \\\cline{2-3} 
& Grad-CAM & .266 \\\hline
\multirow{3}{*}{\shortstack{ISIC \\ Lesions}} & CEM-MAF & -.976 \\\cline{2-3} 
& LIME & -.986   \\\cline{2-3} 
& Grad-CAM  & -.772 \\\hline
\end{tabular}
\label{tab:performance}
\vspace{1mm}
\end{wraptable}
}

A performance comparison of PPs between CEM-MAF, LIME, and Grad-CAM is given in Table 1. CEM-MAF finds a much sparser subset of superpixels than LIME and Grad-CAM and is guaranteed to have the same prediction as the original image. Both LIME and Grad-CAM select features for visual explanations that result in incorrect predictions (low PP accuracy) on color images (particularly on CelebA). A third measure, PP Correlation, measures the benefit of each additional feature by ranking prediction scores after each feature is added (confidence in the prediction should increase) and correlating with expected ranks (perfect correlation here would give -1). Order for LIME was determined by classifier weights while order for Grad-CAM was determined by the corresponding heatmaps. CEM-MAF is best at selecting features that increase confidence on color images, and performs on par with LIME on easier grayscale images. 

The contrastive explanations offered by the PNs offer much more intuition. For example, the first lesion is classified as a Melanoma because it is not rounder, in which case it would be classified as a Nevus (the PN). In fact, Melanomas are known to be asymmetrical due to the quick and irregular growth of cancer cells in comparison to normal cells. In the second column, the Benign Keratosis lesion is classified as such because it is not darker, in which case it would be classified as a Melanoma (the darker center creates the multi-colored lesion associated with Melanoma)
\vspace{-.2cm}
\eat{
\subsection{Value of $L_1$ Regularization in PNs} \label{ss:regularization_l1}
Two key regularizations in the pertinent negative optimization problem (\ref{eqn_per_neg}) maintain that the latent representations of the original image and its PN remain close and that only a few of the attribute functions exhibit changes \eat{(i.e., the PN is the result of modifying only a few features on the original image)}. One might ask whether both regularizations are necessary, and in fact, the framework does allow for including either regularization separately or jointly by setting appropriate penalty values. 

Figure \ref{fig:latent_vs_sparsity} illustrates the usefulness of these regularizations on an image of a young, smiling, male (left image). Regularizing only the latent representation proximity results in a PN that is a young, smiling, female (middle image), from which we can explain that the original image is a male because of the absence of makeup, bangs, and brown hair. The brown hair makes sense because the original image is not defined by any hair color, while female hair color is often easier to detect because females color their hair more often resulting in stronger color \cite{mencolor,womencolor}. Interestingly, three attributes were used to explain the female PN. However, adding sparsity to the number of selected attributes results in the not smiling PN (right image) obtained by solely modifying the cheekbone attribute. The lesson is that both forms of regularization add something to the explanation. Proximity of latent representations keeps the image visually similar (a close inspection shows the male and female having similar facial structure, smile, nose, etc.), while sparsity of modifications can be used to keep the explanation simpler resulting in minimal additions.

\begin{figure}[t]
\begin{center}
  \begin{tabular}{ccc}
   & & \textbf{Latent} \\
    \textbf{Original} & \textbf{Latent} & \textbf{+ Sparsity} \\ \includegraphics[width=.13\textwidth]{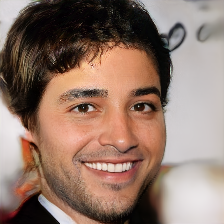} &
    \includegraphics[width=.13\textwidth]{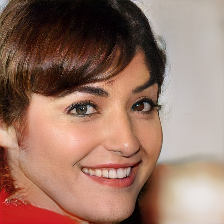} &
    \includegraphics[width=.13\textwidth]{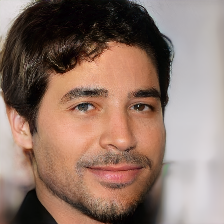} 
        \end{tabular}
  \caption{Illustration of different regularizations to obtain PN. Regularizing only proximity of latent representation explains why the male is not female, while also regularizing attribute sparsity explains why the male is smiling.}
    \label{fig:latent_vs_sparsity}
\end{center}
\end{figure}
}

\subsection{CelebA with Available High-Level Features}
CelebA is a large-scale dataset of celebrity faces annotated with 40 attributes \cite{celebA}. The CelebA experiments explain an 8-class classifier learned from the following binary attributes: Young/Old, Male/Female, Smiling/Not Smiling. We train a Resnet50 \cite{resnet} architecture to classify the CelebA images. We selected the following 11 attributes as our $\{g_i\}$ based on previous studies \cite{dipvae} as well as based on what might be relevant for our class labels: High Cheekbones, Narrow Eyes, Oval Face, Bags Under Eyes, Heavy Makeup, Wearing Lipstick, Bangs, Gray Hair, Brown Hair, Black Hair and Blonde Hair. \eat{Note that this list does not include the attributes that define the classes because an explanation for someone that is smiling which simply says they are smiling would not be useful. Note that the usefulness of CEM-MAF is directly a function of the accuracies of the attribute functions.} See Supplement for details on training these attribute functions and the GAN used.

\subsubsection{CelebA Observations}
Results on six images are exhibited in Figure \ref{fig:celeba}(b). Rows are the same as described for ISIC Lesions. First consider class explanations given by the PPs. Age seems to be captured by patches of skin or hair, sex by patches of skin particular around the lips, and smiling by the presence or absence of the mouth. PPs capture a part of the smile for those smiling, while leave out the mouth for those not smiling. Visually, these explanations are simple (very few selected features) and useful although they require human analysis. In comparison, LIME selects many more features that are relevant to the prediction, and while also useful, requires even more human intervention to explain the classifier. Grad-CAM, which is more useful for discriminative tasks, seems to focus on the mouth and does not always find a part of the image that is positively relevant to the prediction. 

More intuitive explanations are offered by the pertinent negatives in Figure \ref{fig:celeba}. The first PN changes a young, not smiling, male into an young, smiling, male by adding high cheekbones (the mouth is elongated for a closed-mouth smile). Younger age, as predicted by the model, is explained in the second PN by the absence of an oval face; a change in age is represented by a change in face shape. The third PN changes a female into a male, and the female is explained by the absence of a single hair color (in fact, she has black hair with brown highlights) and the presence of bangs. While the presence of bangs is intuitive, it is selected because our constraints of adding features to form PNs can be violated due to enforcing constraints with regularization. The fourth PN again explains the lack of a smile by the absence of high cheekbones, also highly intuitive. The last two PNs explain a male as lacking makeup; notice that both PNs have shinier lips and smoother skin as a result of makeup. 

\eat{Note that the first example where the male was changed to a female in the PN was obtained by regularizing only the latent representation proximity. Our framework allows for different regularizations, and in fact, regularizing sparsity in the number of attributes for this example results in a PN with only a single, rather than three, added attributes classified as a young male that is not smiling (image in Supplement).  \eat{The lesson is that both forms of regularization add something to the explanation. Proximity of latent representations keeps the image visually similar (a close inspection shows the male and female having similar facial structure, smile, nose, etc.), while sparsity of changes can be used to keep the explanation simpler resulting in minimal additions.}}

\begin{figure}[t]
\begin{center}
  \begin{tabular}{cc}
 Increasing Sleeve/Heel & Increasing Cloth/Shoe Fabric\\
   \includegraphics[width=.45\textwidth]{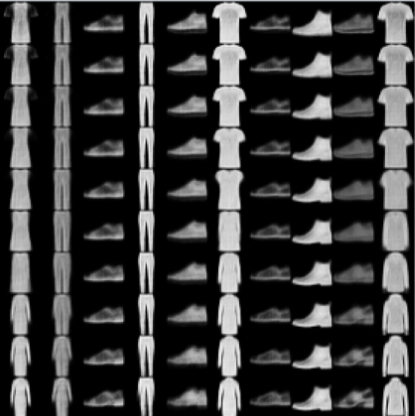} &
    \includegraphics[width=.45\textwidth]{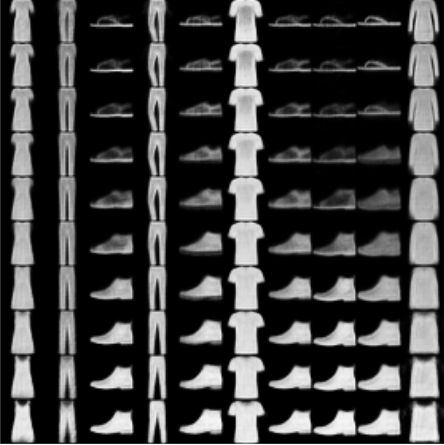}
        \end{tabular}
  \caption{Feature disentanglement of Fashion-MNIST images. Both images represent one-dimensional features that vary in [-3, 3]. As the feature shown in the left figure increases from -3 to 3, a corresponding shirt or shoe would increase in sleeve or heel length, respectively. As the feature shown in the right figure increases from -3 to 3, a corresponding shirt would add central fabric (removing sleeves) or a sandal adds shoe fabric to become a sneaker or boot.}
    \label{fig:disentangled_features_fmnist}
\end{center}
\vskip  -.2in
\end{figure}

\begin{figure}[t]
\centering \small
  \begin{tabu}{|c|c|c|c|c|c|c|}
\hline
\raisebox{0.5ex}{\textbf{\shortstack{Orig. \\ Class}}}
& \raisebox{1.0ex}{\shortstack{T-Shirt}}
& \raisebox{1.0ex}{\shortstack{Pants}}
& \raisebox{1.0ex}{\shortstack{Dress}}
& \raisebox{1.0ex}{\shortstack{Sandal}}
& \raisebox{1.0ex}{\shortstack{Snkr}}
& \raisebox{1.0ex}{\shortstack{Sandal}}
\\ \hline

  \raisebox{3.0ex}{\textbf{Orig.}}
 &\includegraphics[width=.04\textwidth]{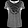}
& \includegraphics[width=.04\textwidth]{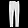}
& \includegraphics[width=.04\textwidth]{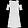}
& \includegraphics[width=.04\textwidth]{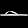}
& \includegraphics[width=.04\textwidth]{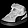}
  & \includegraphics[width=.04\textwidth]{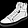}
\\ \hline
 
\raisebox{0.25ex}{\textbf{\shortstack{MAF \\ PN }}}
& \raisebox{1.0ex}{Shirt}
& \raisebox{1.0ex}{Coat}
& \raisebox{1.0ex}{T-Shirt}
& \raisebox{1.0ex}{Snkr}
& \raisebox{0.1ex}{\shortstack{Ankle \\ -boot}}
& \raisebox{0.1ex}{\shortstack{Ankle \\ -boot}}
\\ \hline

\raisebox{0.5ex}{\textbf{\shortstack{MAF \\ PN}}}
&\includegraphics[width=.04\textwidth]{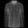}
& \includegraphics[width=.04\textwidth]{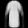}
& \includegraphics[width=.04\textwidth]{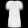}
& \includegraphics[width=.04\textwidth]{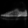}
&\includegraphics[width=.04\textwidth]{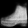}
& \includegraphics[width=.04\textwidth]{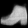}
\\ \hline

 \raisebox{0.5ex}{\textbf{\shortstack{MAF\\PN \\ Expl.}}}
&\raisebox{0.5ex}{\shortstack{+sleeve,\\+fabric}}
&\raisebox{0.5ex}{\shortstack{+sleeve,\\+fabric}}
&\raisebox{3.0ex}{+fabric}
&\raisebox{3.0ex}{+fabric}
&\raisebox{0.5ex}{\shortstack{+heel,\\+fabric}}
&\raisebox{3.0ex}{+heel}
\\ \hline

\raisebox{0.5ex}{\textbf{\shortstack{CEM \\ PN}}}
& \raisebox{1.0ex}{Shirt}
& \raisebox{0.1ex}{\shortstack{Not \\ Found}}
& \raisebox{1.0ex}{Shirt}
& \raisebox{1.0ex}{Snkr}
& \raisebox{1.0ex}{Sandal}
& \raisebox{1.0ex}{Snkr}
\\ \hline

\raisebox{0.5ex}{\textbf{\shortstack{CEM \\ PN}}}
 & \includegraphics[width=.04\textwidth]{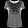}
& \includegraphics[width=.04\textwidth]{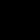}
& \includegraphics[width=.04\textwidth]{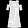}
& \includegraphics[width=.04\textwidth]{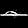}
& \includegraphics[width=.04\textwidth]{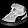}
& \includegraphics[width=.04\textwidth]{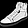}
\\ \hline

\raisebox{0.5ex}{\textbf{\shortstack{MAF \\ PP}}}
 & \includegraphics[width=.04\textwidth]{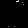}
&\includegraphics[width=.04\textwidth]{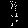}
&\includegraphics[width=.04\textwidth]{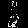}
&\includegraphics[width=.04\textwidth]{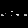}
&\includegraphics[width=.04\textwidth]{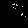}
& \includegraphics[width=.04\textwidth]{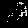}
\\ \hline
 
\raisebox{0.5ex}{\textbf{\shortstack{CEM \\ PP}}}
 & \includegraphics[width=.04\textwidth]{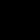}
& \includegraphics[width=.04\textwidth]{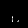}
& \includegraphics[width=.04\textwidth]{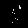}
& \includegraphics[width=.04\textwidth]{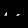}
& \includegraphics[width=.04\textwidth]{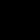}
& \includegraphics[width=.04\textwidth]{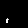}
\\ \hline

\raisebox{3.0ex}{\textbf{LIME}}
 & \includegraphics[width=.04\textwidth]{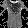}
& \includegraphics[width=.04\textwidth]{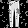}
& \includegraphics[width=.04\textwidth]{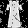}
& \includegraphics[width=.04\textwidth]{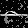}
& \includegraphics[width=.04\textwidth]{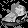}
& \includegraphics[width=.04\textwidth]{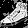}
\\ \hline
 
  \raisebox{1.5ex}{\textbf{\shortstack{Grad \\-CAM}}}
 &\includegraphics[width=.04\textwidth]{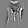}
&\includegraphics[width=.04\textwidth]{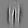}
&\includegraphics[width=.04\textwidth]{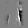}
&\includegraphics[width=.04\textwidth]{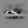}
&\includegraphics[width=.04\textwidth]{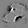}
&\includegraphics[width=.04\textwidth]{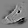}
\\ \hline
    \end{tabu}
  \caption{CEM-MAF examples on Fashion-MNIST. Abbreviations are: Orig. for Original, PN for Pertinent Negative, PP for Pertinent Positive, and Expl. for Explanation. MAF PN/PP compares CEM-MAF against CEM PN/PM. Change from original prediction using CEM-MAF is based on change in disentangled features.}
    \label{fig:fmnist_maf}
\end{figure}

\subsection{Fashion-MNIST with Disentangled Features} 
Fashion-MNIST is a grayscale image dataset of fashion items, akin to the MNIST dataset of handwritten digits commonly used for demonstrating machine learning tasks. Regarding explainability, we demonstrate that this is a more difficult dataset than MNIST, on which the original CEM \cite{CEM} was demonstrated. For our purposes, two datasets are created as subsets of Fashion-MNIST, one for clothes (tee-shirts, trousers, pullovers, dresses, coats, and shirts) and one for shoes (sandals, sneakers, and ankleboots). Our setup processes each item as a $28 \times 28$ pixel grayscale image. As with the ISIC Lesions dataset, a VAE was trained to learn the Fashion-MNIST manifold, and we again learn the features using DIP-VAE due to a lack of annotated features. For each of these datasets, we explain a CNN model with two convolutional layers followed by two dense layers. The architectures for training the CNN and DIP-VAE models on Fashion-MNIST are detailed in the Supplement.

Based on what might be relevant to the clothes and shoes classes, we use four dimensions from the latent codes corresponding to sleeve length, shoe mass, heel length and waist size. Given these attributes, we learn two classifiers, one with six classes (clothes) and one with three classes (shoes). Figure \ref{fig:disentangled_features_fmnist} visualizes two features, one corresponding to increasing sleeve/heel length, and the other corresponding to increasing the amount of fabric.

\subsubsection{Fashion-MNIST Observations}
Results on three clothes images and three shoe images are shown in Figure \ref{fig:fmnist_maf}. In addition to LIME and Grad-CAM, comparisons are made with CEM \cite{CEM}, which as already discussed, is only amenable to grayscale image datasets such as this one, i.e., it cannot be used to generate explanations for either the ISIC Lesions or CelebA datasets in the previous sections. We here demonstrate another benefit of CEM-MAF over CEM, besides applicability on color images, is better results on datasets on which pertinent negatives are not easily obtainable. In order to make a fair comparison to CEM we do not segment the images (as done with the color datasets), but rather do feature selection pixel-wise since CEM cannot make use of a segmentation.

Consider first the PPs. Both CEM-MAF and CEM offer mostly sparse explanations and do not give visually intuitive explanations. The CEM-MAF explanations for shoes use more pixels than CEM to highlight the soles of sandals versus sneakers. Both LIME and Grad-CAM, by selecting many more relevant pixels, offer a much more visually appealing explanation. However, these explanations simply imply, for example, that a T-Shirt in the first column is a T-Shirt because it looks like a T-Shirt. These two datasets (clothes and shoes) are examples where the present features do not offer an intuitive explanation.

Rather, contrastive explanations (relative to other items) about what is absent leads to the most intuitive explanations for these images. That same T-Shirt is a T-shirt because it does not have longer sleeves and additional fabric material. The pair of pants is predicted to be a pair of pants because, if it had sleeves and additional fabric between the legs, it would be predicted a coat. This specific example was too difficult for CEM to even find a contrastive image with a reasonable sparsity on adding pixels. CEM usually finds a contrastive example by flipping a few pixels, which to the human eye, is mostly indifferent from the original image. For example, the CEM-MAF PN for a sneaker is an ankleboot made by adding a heel and additional fabric to the shoe, whereas the CEM PN for a sneaker is a sandal and it is not clear why. Handwritten digits have much more structure, and for this reason, CEM is much more successful and useful on MNIST as illustrated in \cite{CEM}.
\vspace{-.05in}

\section{User Study}
In order to further illustrate the usefulness of, and preference for, contrastive explanations, we conducted a user study comparing CEM-MAF, LIME, and Grad-CAM. Our study assesses two questions: are the explanations useful by themselves and whether or not an additional explanation would add to the usefulness of an initial explanation. This is in line with a user study in \cite{grace} that compares LIME with a recent contrastive explanation method for tabular data called GRACE by testing if "GRACE is more intuitive and friendly" and if "GRACE is more comprehensible" relative to LIME.

Regarding image classification, LIME and Grad-CAM clearly have usefulness as explanation methods, evidenced by previous literature. Contrastive explanations offer different information that could be viewed as complimentary to the visual explanations given by the other methods. Pertinent positives, while also visual, have a specific meaning in highlighting a minimally sufficient part of the image to obtain a classifier's prediction, while LIME highlights significant features using local linear classifiers and Grad-CAM highlights features that attribute maximum change to the prediction. Each method provides different information. Pertinent negatives illuminate a completely different aspect that is known to be intuitive to humans \cite{Timcontras}, and importantly does not require humans to interpret what is learned since the method outputs exactly what modifications are necessary to change the prediction.

The study was conducted on Amazon Mechanical Turk using CelebA images. A single trial of the study proceeds as follows. An original image is displayed along with the model's prediction in a sentence, e.g.,``The system thinks the image shows a young male who is smiling." Under the image, one of the three explanation methods, titled ``Explanation 1" for anonymity, is displayed with a brief description of the explanation, e.g. ``The colored pixels highlight the parts of the original image that may have contributed significantly to the system's recognition result." The question, ``Is the explanation useful?", is posed next to Explanation 1 along with five choices ranging from ``I don't understand the explanation" to ``I understand the explanation and it helps me completely understand what features the system used to recognize a young, male, who is smiling" (the words \eat{for the prediction} change per image).

\begin{figure}[t]
\begin{center}
  \begin{tabular}{c}
  CEM-MAF vs Grad-CAM vs LIME\\
   \includegraphics[width=.6\textwidth]{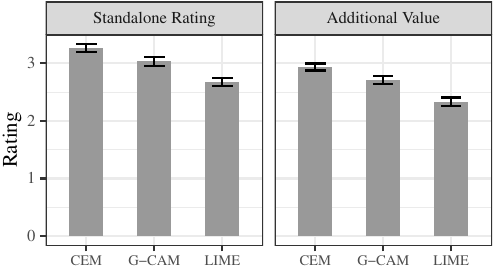} \\ \\
   PN vs PP vs Grad-CAM vs LIME\\
   \includegraphics[width=.6\textwidth]{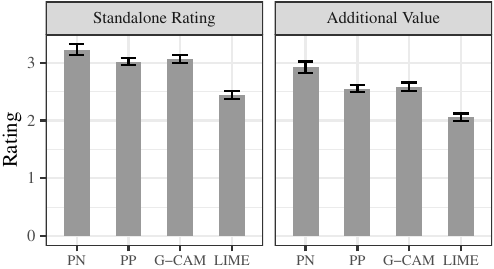} \\
        \end{tabular}
  \caption{User study results. Top figures compare CEM-MAF with Grad-CAM (G-CAM) and LIME, where figure (left) rates the usefulness of explanations while figure (right) rates additional value given one of the other explanations. Bottom figures show similar ratings but separates CEM-MAF into PN and PP. Error bars show standard error.}
    \label{fig:human_study}
\end{center}
\vskip  -.2in
\end{figure}

After answering the first question, another image, titled ``Explanation 2", is displayed under the first explanation, again with a short blurb about the explanation result. If the first explanation was LIME, the second explanation would be CEM-MAF or Grad-CAM. The question posed next to this explanation is ``How much does adding the second explanation to the first improve your understanding of what features the system used to recognize a young male who is smiling?" There are five possible responses ranging from ``None" to ``Completes my understanding." In a second survey, we ask the same questions for four methods, where we separate CEM-MAF and pose the PP and PN as two different explanations.\eat{ A screenshot of a single trial can be found at the end of the Supplement.}

Each user in the study is required to spend 30 seconds reading instructions that gives examples before proceeding through 24 trials as described above. Each of the three methods is assigned to both explanations 1 and 2 such that a user sees each of the six permutations four times (the second survey with four explanations has twelve permutations that are each seen twice). Each trial draws a random image from a batch of 76 images for which we have ran all methods. As LIME requires tuning, we restrict LIME to use the same number of superpixels as used in the pertinent positives for an apples-to-apples comparison. Regarding demographics, we attempt to run the survey over a population that is more likely to use the methods of this paper by restricting job function of participants to Information Technology. Each survey was taken by 100 participants (200 total). 

This setup was settled on after consulting various works from the explainable AI evaluation literature (e.g., \cite{hoffman_metrics_2019, bucinca2020, poursabzi2018manipulating, mohseni2018survey, carvalho_machine_2019, grace}). Our primary reasons for this approach are: 1) evaluating performance on a decision-making task would be insufficient because users can easily classify facial expressions without the model's help, 2) a proxy task, where a participant is asked to adjust the features of a sample in a post-explanation decision scenario as in \cite{grace}, is amenable to tabular but not image data, and 3) a proxy task, where a participant is asked to guess what the model will predict, is likely to favor methods that show more image pixels rather than just sufficient parts; a similar phenomenon was observed by \cite{bucinca2020}. Hence, we were left with subjective evaluation, where our two-step questions are meant to essentially evaluate the necessity of information provided by each explanation. We believe this helps alleviate key issues associated with subjective evaluation.

Results are displayed in Figure \ref{fig:human_study} with error bars showing one standard error. The standalone ratings judge individual usefulness of all explanation methods. CEM-MAF has a clear, albeit, slight advantage in terms of usefulness over other methods, and when broken down, it is apparent that the usefulness is driven more by the PN than the PP, which is on par with Grad-CAM. The direct explanation that requires no human intervention to interpret the explanation seems to be most useful (i.e., PNs directly output via text what the explanation is whereas other methods require a human to interpret what is understood visually). Given one explanation, the additional value of offering a second explanation follows a similar pattern to the standalone ratings. This means that users given a Grad-CAM or LIME explanation found the contrastive explanation to offer more additional understanding than if they were first given CEM-MAF followed by Grad-CAM or LIME. Note these results are averaged, for example, with CEM-MAF as the second explanation, over all trials with either Grad-CAM or LIME as the first explanation. P-values for pairwise t-tests comparing CEM-MAF to Grad-CAM in Figure \ref{fig:human_study} (top) are 0.0018 and 0.0006 for Standalone and Additional, respectively, so differences between CEM-MAF and both other methods are highly significant.

While CEM-MAF was observed as the most useful explanation, comments made at the end of the survey suggest what has been mentioned earlier: each of the different explanation methods have value. \eat{Such comments included ``The explanations were straightforward" and ``I thought they were pretty accurate overall and helped in the decision making process". }Other comments suggested that there is still much variability in the success of explanation methods, e.g., \eat{``I thought they were very unique and amazing to see. I thought the explanations were usually a bit confused, but some were very advanced" and }``The explanation made sense to me, though I think some of them were lacking in complexity and detail." Such comments also point out where contrastive explanations can have major impact. For example, since PP's dictate when a classifier requires little information to make its prediction (i.e., require ``little complexity and detail"), they can be used to locate classifiers that should be reassessed prior to deployment.

\eat{
While it may be clear that contrastive explanations offer much intuition as to why classifiers output their particular predictions, we conducted a study in order to illustrate that humans find a strong preference to having explicit explanations (as dictated by the pertinent negatives and which attributes were added). Such explanations are preferable to visual explanations where the human must determine the intuition themselves. Our study sent Figure \ref{fig:celeba} (a) to 25 individuals and asked which method among CEM-MAF, LIME, and Grad-CAM were most useful in explaining the original predictions and the votes were 20, 1, 3, respectively, with 1 response saying no methods were useful. 

Feedback was given as well. The most interesting and common comment was that neither PPs, LIME, nor Grad-CAM were useful in explaining predictions for this task, and that the only relevant information given was the PN attribute descriptions. This is significant because no other explanation method offers human understandable descriptions.  All other methods require human intervention, typically to learn something from a visual explanation. In CEM-MAF, the only human intervention may be in the preproccessing phase for interpreting latent features learned in DIP-VAE, after which, the users have no need for interpreting because CEM-MAF directly outputs the descriptions to them.
}

\eat{
\section*{Ethical and Societal Impact}
Our work extends previous tools developed for contrastive explanations to color images where the concept of adding features is ill-defined. Many important applications require the use of color images such as the lesion classification problem illustrated in the paper. Automating the detection of disease has obvious potential benefits ranging from early detection (because one doesn't have to wait for their next doctor appointment) to financial (preventing unnecessary doctor appointments and examinations). AI is already being used in such domains, but the models are typically black-box and building trust in the predictions is an important problem where providing contrastive explanations can help. The contrastive explanations developed in this work extend the previous toolbox to explaining models that act on data with rich (hidden) structures. They provide more intuitive explanations based on these structures, where previous methods were only amenable to models on simpler black/white images or tabular data.

Since AI mostly serves to complement experts in these domains, building incorrect trust is a risk, particularly with models that do not perform well. For example, we do not want to build trust in a poor performing model that informs a doctor, but rather make sure the doctor knows not to trust the model because of the explanations. Furthermore, even for a good model, as we've noted regarding the user study, different explanation methods offer different pieces of information. It is possible (or even likely) that no explanation method is, by itself, complete, particularly in the case of explaining a black-box model; there is great risk in many applications when acting on partial information. Take the HELOC example - an applicant might admit more risk knowing he/she will still be in good standing, but perhaps it is not easy to undo the risk after it has been taken. It is imperative that the applicant understand what actions would need to be taken to regain the previous higher level of standing, should that ever be required. In general, this work has broad appeal because it does not require the dataset to be annotated. The framework can provide explanations using known high-level features or learned latent features.
}
\vspace{-.05in}
\section{Discussion}
In this paper, we leveraged high-level latent features that were readily available (viz. CelebA) as well as used generated ones (viz. ISIC Lesions and Fashion-MNIST) based on disentanglement to produce contrastive explanations for the predictions of supervised models. \eat{As mentioned before, one could also learn interpretable features in a supervised manner as done in \cite{tcav} which we could also use to generate such explanations. Regarding learning the data manifold, it is important to note that we do not need a global data manifold to create explanations; a good \emph{local} data representation around the image should be sufficient. Thus, for datasets where building an accurate global representation may be hard, a good local representation (viz. using conditional GANs) should make it possible to generate high quality contrastive explanations.} Several future directions are motivated by the user study. Extracting text to describe highlighted aspects of visual explanations (PP, LIME, and Grad-CAM) could better help their understanding by humans. As previously mentioned, an important feature of contrastive explanations is that they can be used to identify classifiers that should not be trusted. PPs composed of two superpixels in our CelebA experiments could identify examples where the classifier requires little relevant information, and hence might not be operating in a meaningful way. The study reinforced this concept because users found visual explanations with few superpixels to be less useful. Studying this relationship between human trust and contrastive explanations is an interesting direction.  

In summary, we have proposed a method to create contrastive explanations for image data with rich structure. The key novelty over previous work is how we define addition, which leads to realistic images where added information is easy to interpret (e.g., added makeup). Our results showcase that using pertinent negatives might be the preferred form of explanation where it is not clear why a certain entity is what it is in isolation (e.g. based on relevance or pertinent positives), but could be explained more crisply by contrasting it to another entity that closely resembles it (viz. adding high cheekbones to a face makes it look smiling).

\eat{Several future directions are motivated by the user study. Extracting text to describe highlighted aspects of visual explanations (PP, LIME, and Grad-CAM) could better help their understanding by humans. Regarding human studies, there is much more to understand about human biases with visual explanations, as we observed that people tend to ``understand'' explanations with more superpixels regardless of the lack of highlighting key features. Another aspect to study is population bias - broadening the population found participants that initially found most explanations to be useless.}

In the future, we would like to study new regularizations that better enforce our constraints and produce interpretable additions as we have done here. We would also like to investigate robustness of our explanations to errors in the attribute functions. One possibility to make our explanations more robust is to jointly optimize the attributes functions along with the contrastive explanation objective. This may be more computationally expensive, but could potentially provide better quality explanations. Another independent direction could be to generalize our methods to work where we have limited access to the classification model (viz. only query access). In such a scenario, creating contrastive explanations for images that are also qualitatively good might be a challenge.


\bibliography{CEMColoredImages_arXiv}
\bibliographystyle{plain}

\clearpage
\section*{Supplement}
\setcounter{section}{0}
We present here additional information relevant to replication and understanding of the experiments. The first section details the hyperparameters used in all experiments to compute pertinent positives and pertinent negatives. The following three sections offer additional information on the three datasets, CelebA, ISIC Lesions, and Fasion-MNIST, relevant to training models and learning disentangled features for the ISIC Lesions and Fashion-MNIST datasets.

\section{Hyperparameter selection for PN and PP}
Finding PNs is done by solving (\ref{eqn_per_neg}) which has hyperparameters $\kappa, \gamma, \beta,\eta, \mu, c$. The confidence parameter $\kappa$ is the user's choice. We experimented with $\kappa\in\{0.02, 0.05, 0.10\}$ and report results with $\kappa=0.02, 0.05, 0.05$ for celebA, ISIC Lesions and Fashion-MNIST, respectively, corresponding to a PN being predicted with 0.02, 0.05, and 0.05 more confidence than the prediction on the original class. We experimented with $\gamma\in\{1, 100\}$ and report results with $\gamma=100$ which better enforces the constraint of only adding attributes to a PN. Attribute sparsity parameter $\beta$ was varied in $\{.001, .01, .1, 1, 10, 100$ to find sufficient sparsity in attributes and best values for celebA, ISIC Lesions and Fashion-MNIST were found to be 100, 1, and .01, respectively. Sufficient sparsity in attributes was obtained with these values of $\beta$ but increasing $\beta$ could be done to allow for more attributes if desired. The two hyperparameters $\eta=1.0$ and $\nu=1.0$ were fixed; results with these values were deemed realistic so there was no need to further tune them. The last hyperparameter $c$ was selected via the following search: Start with $c=1.0$ and multiply $c$ by $10$ if no PN found after 1000 iterations of subgradient descent, and divided by 2 if PN found. Then run the next 1000 iterations and update $c$. This search on $c$ was repeated 9 times, meaning a total of $1000\times 9 =9000$ iterations of subgradient descent were run with all other hyperparameters fixed.  

Finding PPs is done by solving (\ref{eqn_per_pos}) which has hyperparameters $\kappa, \gamma, \beta, c$. Again, we experimented with $\kappa\in\{0.02, 0.05, 0.10\}$ and report results with $\kappa=0.02, 0.05, 0.05$ for celebA, ISIC Lesions and Fashion-MNIST, respectively, corrresponding to confidence in a PP being no less than 0.02, 0.05, and 0.05 within the confidence of the original prediction. We experimented with $\gamma\in\{1, 100\}$ and report results with $\gamma=100$ for the same reason as for PNs. Mask sparsity parameter $\beta$ was varied in $\{.001, .01, .1, 1, 10, 100$ to find sufficient sparsity in attributes and best values for celebA, ISIC Lesions and Fashion-MNIST were found to be 0.01, 0.01, and .001, respectively. Higher values of $\beta$ were too strong and did not find PPs with such high sparsity (usually allowing no selection). The same search on $c$ as described for PNs above was done for PPs.

\section{Additional CelebA Information}
We here discuss how attribute classifiers were trained for CelebA and describe the GAN used for generation. CelebA datasets are available at \url{http://mmlab.ie.cuhk.edu.hk/projects/CelebA.html}.

\subsection{Training attribute classifiers for CelebA}
For each of the 11 selected binary attributes, a CNN with four convolutional layers followed by a single dense layer was trained on 10000 CelebA images with Tensorflow's SGD optimizer with Nesterov using learning rate=0.001, decay=1e-6, and momentum=0.9. for 250 epochs. \eat{Accuracies of each classifiers are given in Table \ref{table:celeba_attributes}.}
\eat{
\begin{table}[b]
\vspace{8mm}
\begin{center}
\caption{CelebA binary attribute classifer accuracies on 1000 test images}
\begin{tabular}{|c|c|}
\hline
\textbf{Attribute}  & \textbf{Accuracy}  \\ \hline
High Cheekbones   & 84.5\% \\ \hline
Narrow Eyes & 78.8\% \\ \hline
Oval Face & 63.7\% \\ \hline
Bags Under Eyes & 79.3\%\\ \hline
Heavy Makeup& 87.9\%\\ \hline
Wearing Lipstick& 90.6\%\\ \hline
Bangs & 93.0\% \\ \hline
Gray Hair & 93.5\%\\ \hline
Brown Hair & 76.3\%\\ \hline
Black Hair & 86.5\%\\ \hline
Blonde Hair &  93.1\%\\ \hline
\end{tabular}
\label{table:celeba_attributes}
\end{center}
\end{table}
}
\subsection{GAN Information}
Our setup processes each face as a $224 \times 224$ pixel colored image. A GAN was trained over the CelebA dataset in order to generate new images that lie in the same distribution of images as CelebA. 
Specifically, we use the pretrained progressive GAN available at \url{https://github.com/tkarras/progressive_growing_of_gans} \eat{in \cite{karras2017progressive}} to approximate the data manifold of the CelebA dataset. The progressive training technique is able to grow both the generator and discriminator from low to high resolution, generating realistic human face images at different resolutions.

\section{Additional ISIC Lesions Information}
We here discuss what the different types of lesions in the ISIC dataset are and how disentangled features were learned for ISIC lesions. The ISIS Lesions dataset is available at \url{https://challenge2018.isic-archive.com/}. 

\subsection{Lesion Descriptions}
The ISIC dataset is composed of dermoscopic images of lesions classified into the following seven categories: Melanoma, Melanocytic Nevus, Basal Cell Carcinoma, Actinic Keratosis, Benign Keratosis, Dermatofibroma, and Vascular Lesions. Information about these categories is available at \url{https://challenge2018.isic-archive.com/} but we give some key information about these lesions below.  All description were obtained from \url{www.healthline.com/health}. 

\noindent\textbf{Melanoma}: Cancerous mole, asymmetrical due to cancer cells growing more quickly and irregularly than normal cells, fuzzy border, different shades of same color or splotches of different color (but not one solid color). Many different tests to diagnose the stage of cancer.

\noindent\textbf{Melanocytic Nevus}: Smooth, round or oval-shaped mole.  Usually raised, single or multi-colored, might have hairs growing out. Pink, tan, or brown. Might need biopsy to rule out skin cancer.

\noindent\textbf{Basal Cell Carcinoma}: At least 5 types, pigmented (brown, blue, or black lesion with translucent and raised border), superficial (reddish patch on the skin, often flat and scaly, continues to grow and often has a raised edge), nonulcerative (bump on the skin that is white, skin-colored, or pink, often translucent, with blood vessels underneath that are visible, most common type of BCC), morpheaform (least common type of BCC, typically resembles a scarlike lesion with a white and waxy appearance and no defined border), basosquamous (extremely rare).

\noindent\textbf{Actinic Keratosis}: Typically flat, reddish, scale-like patch often larger than one inch. Can start as firm, elevated bump, or lump. Skin patch can be brown, tan, gray, or pink. Visual diagnosis, but skin biopsy necessary to rule out change to cancer.

\noindent\textbf{Benign Keratosis}: Starts as small, rough, area, usually round or oval-shaped. Usually brown, but can be yellow, white, or black. Usually diagnosed by eye. 

\noindent\textbf{Dermatofibroma}: Small, round noncancerous growth, different layers. Usually firm to the touch. Usually a visual diagnosis.

\noindent\textbf{Vascular Lesions}: *Too many types to list.

\subsection{Learning Disentangled Features for ISIC Lesions}
Our setup processes each item as a $128 \times 128$ pixel color image. As with many real-world scenarios, ISIC Lesion samples come without any supervision about the generative factors or attributes. For such data, we can rely on latent generative models such as VAE that aim to maximize the likelihood of generating new examples that match the observed data. VAE models have a natural inference mechanism baked in and thus allow principled enhancement in the learning objective to encourage disentanglement in the latent space. For inferring disentangled factors, inferred prior or expected variational posterior should be factorizable along its dimensions. We use a recent variant of VAE called DIP-VAE \cite{dipvae} that encourages disentanglement by explicitly matching the inferred aggregated posterior to the prior distribution. This is done by matching the covariance of the two distributions which amounts to decorrelating the dimensions of the inferred prior. Table \ref{table:DIP_VAE_isic} details the architecture for training DIP-VAE on ISIC Lesions.

\begin{table}[t]
\begin{center}
\caption{Details of the model architecture used for training DIP-VAE \cite{dipvae} on ISIC Lesions.}
\begin{tabular}{cl}
\hline
\textbf{}  & \textbf{Architecture}  \\ \hline
Input & 16384 (flattened 128x128x1)\\
Encoder & Conv 32x4x4 (stride 2), 32x4x4 (stride 2), \\
& 64x4x4 (stride 2), 64x4x4 (stride 2), FC 256. \\
& ReLU activation \\
Latents & 32 \\
Decoder & Deconv reverse of encoder. ReLU activation. \\ & Gaussian \\
\hline
\end{tabular}
\label{table:DIP_VAE_isic}
\end{center}
\end{table}

\section{Additional Fashion-MNIST Information}
Fashion-MNIST is a large-scale image dataset of various fashion items (e.g., coats, trousers, sandals). We here discuss how disentangled features were learned
for Fashion-MNIST, how classifiers were trained, and pro-
vide additional examples of CEM-MAF. Fashion-MNIST
datasets are available at https://github.com/zalandoresearch/
fashion-mnist. 

\subsubsection{Learning Disentangled Features for Fashion-MNIST}
Two data -sets are created as subsets of Fashion-MNIST, one for clothes (tee-shirts, trousers, pullovers, dresses, coats, and shirts) and one for shoes (sandals, sneakers, and ankleboots). Our setup processes each item as a $28 \times 28$ pixel grayscale image. Again, a VAE was trained to learn the Fashion-MNIST manifold, and we again learn the features using DIP-VAE due to a lack of annotated features. Table \ref{table:DIP_VAE_fmnist} details the architecture for training DIP-VAE on Fashion-MNIST.

Based on what might be relevant to the clothes and shoes classes, we use four dimensions from the latent codes  corresponding to sleeve length, shoe mass, heel length and waist size.  Given these attributes, we learn two classifiers, one with six classes (clothes) and one with three classes (shoes).

\begin{table}[t]
\begin{center}
\caption{Details of the model architecture used for training DIP-VAE \cite{dipvae} on Fashion-MNIST.}
\begin{tabular}{cl}
\hline
\textbf{}  & \textbf{Architecture}  \\ \hline
Input & 784 (flattened 28x28x1)\\
Encoder & FC 1200, 1200. ReLU activation.\\
Latents & 16 \\
Decoder & FC 1200, 1200, 1200, 784. ReLU \\
& activation.\\
Optimizer & Adam (lr = 1e-4) with mse loss \\
\hline
\end{tabular}
\label{table:DIP_VAE_fmnist}
\end{center}
\end{table}

\begin{table}[t]
\begin{center}
\caption{Details of the model architecture used for training classifiers on Fashion-MNIST.}
\begin{tabular}{cl}
\hline
\textbf{}  & \textbf{Architecture}  \\ \hline
Input & 28x28x1 \\
Shoes Classifier & Conv (5,5,32), MaxPool(2,2), \\ & Conv(5,5,64), MaxPool(2,2), \\
& Flatten, FC 1024, \\ 
& Dropout (rate=0.4), FC 3, \\
& Relu activation.\\
Clothes Classifier & Conv (5,5,32), MaxPool(2,2), \\ 
& Conv(5,5,64), MaxPool(2,2), \\
& Flatten, FC 1024, \\
& Dropout (rate=0.4), FC 6, \\
& Relu activation.\\
Optimizer & SGD (lr = 1e-3) with cross entropy \\
&loss \\
\hline
\end{tabular}
\label{table:fmnist_classifiers}
\end{center}
\end{table}

\subsection{Training Fashion-MNIST classifiers}
Two datasets are created as subsets of Fashion-MNIST, one for clothes (tee-shirts, trousers, pullovers, dresses, coats, and shirts) and one for shoes (sandals, sneakers, and ankleboots). We train CNN models for each of these subsets with two convolutional layers followed by two dense layer to classify corresponding images from original Fashion-MNIST dataset. See Table \ref{table:fmnist_classifiers} for training details.

\end{document}